\documentclass[twoside]{article}
\usepackage[accepted]{aistats2018}
\usepackage{textcomp, libertine}
\usepackage[utf8]{inputenc} % allow utf-8 input
\usepackage[T1]{fontenc}    % use 8-bit T1 fonts
\PassOptionsToPackage{hyphens}{url}\usepackage{hyperref}
\usepackage{booktabs}       % professional-quality tables
\usepackage{nicefrac}       % compact symbols for 1/2, etc.

\usepackage{graphicx} % more modern
\usepackage{subfigure} 

\usepackage{natbib}

\usepackage{lipsum}

\newcommand\blfootnote[1]{%
  \begingroup
  \renewcommand\thefootnote{}\footnote{#1}%
  \addtocounter{footnote}{-1}%
  \endgroup
}

\usepackage{amsmath}
\usepackage{bm}
\usepackage{amsfonts}
\usepackage{amssymb}
\usepackage{caption}
\DeclareMathOperator{\Tr}{Tr}

\let\oldhref\href
\renewcommand{\href}[2]{\oldhref{#1}{\hbox{#2}}}

\usepackage{algpseudocode,algorithm,algorithmicx}

\algdef{SE}[DOWHILE]{Do}{doWhile}{\algorithmicdo}[1]{\algorithmicwhile\ #1}%

\usepackage[letterpaper, pass]{geometry}
\begin{document}

\twocolumn[

\aistatstitle{Differentially Private Regression with Gaussian Processes}

\aistatsauthor{ Michael T. Smith${}^*$ \And Mauricio A. \'{A}lvarez \And Max Zwiessele \And Neil D. Lawrence }

\aistatsaddress{Department of Computer Science, University of Sheffield} ]

%\aistatsauthor{ Anonymous Authors }
%\aistatsaddress{ Anonymous Institution } ]

\title{Differentially Private Regression with Gaussian Processes}

\begin{abstract} 
A major challenge for machine learning is increasing the availability of data while respecting the privacy of individuals. Here we combine the provable privacy guarantees of the differential privacy framework with the flexibility of Gaussian processes (GPs). We propose a method using GPs to provide differentially private (DP) regression. We then improve this method by crafting the DP noise covariance structure to efficiently protect the training data, while minimising the scale of the added noise. We find that this cloaking method achieves the greatest accuracy, while still providing privacy guarantees, and offers practical DP for regression over multi-dimensional inputs. Together these methods provide a starter toolkit for combining differential privacy and GPs.
\end{abstract} 

\section{Introduction}

As machine learning algorithms are applied to an increasing range of personal data types, interest is increasing in mechanisms that allow individuals to retain their privacy while the wider population can benefit from inferences drawn through assimilation of data. Simple `anonymisation' through removing names and addresses has been found to be insufficient \citep{sweeney1997weaving, ganta2008composition}. Instead, randomisation-based privacy methods (such as differential privacy, DP) provide provable protection against such attacks. 

In this paper we investigate corrupting a Gaussian process's (GP's) fit to the data in order to make aspects of the training data private. Importantly this paper addresses the problem of making the training \emph{outputs} ($\bm{y}$) of GP regression private, not its inputs. To motivate this, consider inference over census data. The inputs to our GP are the locations of every household in the country during a census. Note that the presence of a residential property need not be private. The values associated with that location (e.g. the person's religion or income) are private. Thus we can release the locations of the census households (in X) but protect the census answers (in $\bm{y}$).
%It should be noted that adding a new person will still reveal their location (if side information allows us to know who the additional person is) but this we argue is irrelevant in a census database, in which there are many location changes between the infrequent census surveys.
A second example is from a project in which the method has already been applied, analysing road collision data in Kampala, Uganda. Collision times and locations (which are public record) are entered in X, with the vehicles involved, the ages and genders of the victims, kept private (in $\bm{y}$). We are able to make differentially private inference around (for example) the times/places where children are most likely to be involved in collisions, providing useful insights, while ensuring that information about the victims is kept private.

We approach the problem of applying DP to GPs by finding a bound on the scale of changes to the GP's posterior mean function, in response to perturbations in the training outputs. We then use the results from \citet{hall2013differential} to add appropriate Gaussian DP noise (Section \ref{noiseoutputs}). We find however that the added DP noise for this initial method is too large for many problems. To ameliorate this we consider the situation in which we know \emph{a priori} the locations of the test points, and thus can reason about the specific correlation structure in the predictions for given perturbations in the training outputs (Section \ref{cloaking}). Prior knowledge of the query is not unusual in methods for applying DP. Assuming the Gaussian mechanism is used to provide the DP noise, we are able to find the optimal noise covariance to protect training outputs. Finally we compare this strategy for inducing privacy with a DP query using the Laplace mechanism on bin means \citep[section 3.4]{dwork2014algorithmic}, and show that it provides greater accuracy for a given privacy guarantee for our example dataset.
\blfootnote{${}^*$corresponding author: m.t.smith@sheffield.ac.uk}
It is worth emphasising that we can still release the GP's covariance structure (as this only depends on the input locations, which we assume to be public) and the scale of the DP added noise. Thus the user is able to account for the uncertainty in the result. This paper combines the ubiquity of GP regression with the rigorous privacy guarantees offered by DP. This allows us to build a toolkit for applying DP to a wide array of problems amenable to GP regression.
%
%\vspace{-2mm}
\subsection*{Related Work}
\vspace{-2mm}
A DP algorithm\citep{dwork2006calibrating, dwork2014algorithmic} allows privacy preserving queries to be performed by adding noise to the result, to mask the influence of individual data. This perturbation can be added at any of three stages in the learning process \citep{berlioz2015applying}: to the (i) training data, prior to its use in the algorithm, (ii) to components of the calculation (such as to the gradients or objective) or (iii) to the results of the algorithm. In this paper we focus on (iii) (adding the DP noise to the \emph{predictions} in order to make aspects of the training data private).
Considerable research has investigated the second of these options, in particular fitting parameters using an objective function which has been perturbed to render it differentially private \citep[e.g.][]{chaudhuri2011differentially,zhang2012functional} with respect to the training data, or more recently, \citet{song2013stochastic} described how one might perform stochastic gradient descent with DP updates. Some attention has also been paid to non-parametric models, such as histograms \citep{wasserman2010statistical} and other density estimators, such as the method described in \citet{hall2013differential} which performs kernel density estimation (there are also parametric DP density estimators, such as those described by \citet{wu2016differentially} who use Gaussian mixtures to model density). For regression, besides using a perturbed objective function, one can also use the subsample-and-aggregate framework, as used by \citet[section 7]{dwork2009differential}, effectively protecting the parametric results of the regression. \citet{heikkila2017differentially} use a similar idea for fitting parameters in a distributed, DP manner.
%The framework described in \citet{dimitrakakis2017differential} relies on smoothness assumptions which would...?

There are very few methods to perform DP non-parametric regression. To conduct a comparison we chose a binning method in which we make the data private by manipulating the bin means (using the Laplace mechanism)
%\footnote{We used Laplace noise as it causes a lower RMSE than the Gaussian noise alternative (with $\delta=0.01$).}
as other methods were not appropriate. For example, \citet{chaudhuri2011differentially}, \citet{rubinstein2009learning} and \citet{song2013stochastic} were for classification, while \citet{zhang2012functional} was for parametric models and only considered linear (and logistic) regression. It may be possible to extend their work, but this would be beyond the scope of the paper. \citet{wasserman2010statistical} use histogram queries. \citet{machanavajjhala2008privacy} use the less strict `indistinguishability' definition. In summary, there is a dearth of methods for performing non-parametric differentially private regression. In particular there is an absence of research applying differential privacy to Gaussian processes (in \citet{hall2013differential} they make use of GP's properties to provide DP to functions and vectors, but do not do the converse, making the GP's predictions private).
\vspace{-2mm}
\subsection*{Differential Privacy}
\vspace{-2mm}
Briefly we reiterate the definition of differential privacy, from \citet{dwork2014algorithmic}. To query a database in a differentially private manner, a randomised algorithm $R$ is $(\varepsilon, \delta)$-differentially private if, for all possible query outputs $m$ and for all neighbouring databases $D$ and $D^\prime$ (i.e. databases which only differ by one row),
$$
P\Big( R(D) \in m \Big) \leq e^{\varepsilon} P\Big( R(D^\prime) \in m \Big) + \delta. \label{dpdef}
$$
This says that we want each output value to be almost equally likely regardless of the value of one row: we do not want one query to give an attacker strong evidence for a particular row's value. $\varepsilon$ puts a bound on how much privacy is lost by the query, with a smaller $\varepsilon$ meaning more privacy. $\delta$ says this inequality only holds with probability $1-\delta$.
\section{Applying Differential Privacy to a Gaussian Process}
\label{noiseoutputs}
The challenge is as follows; we have a dataset in which some variables (the inputs, $\bm{X}$) are public, for example the latitude and longitude of all homes in a country. We also have a variable we want to keep secret ($\bm{y}$, e.g. income). We want to allow people to make a prediction about this variable at a new location, while still ensuring that the dataset's secret variables remain private. In this section we fit a standard GP model to a dataset and calculate the bound on the scale of the perturbation we need to add to the posterior mean to provide a DP guarantee on the training outputs.
%
%\subsection{Differential Privacy for Functions}

\citet{hall2013differential} extended DP to functions. Consider a function, $f$, that we want to evaluate (with privacy guarantees). If the family of functions from which this function is sampled lies in a reproducing kernel Hilbert space (RKHS) then one can consider the function as a point in the RKHS. We consider another function, $f^\prime$, that has been generated using identical data except for the perturbation of one row. The distance, $||f - f^\prime||$, between these points is bounded by the sensitivity, $\Delta$. The norm is defined to be $|| g || = \sqrt{\langle g,g \rangle_{H}}$. Specifically the sensitivity is written $\Delta \geq \text{sup}_{D \sim D^\prime} || f_D - f_{D^\prime} ||_H$.
\citet{hall2013differential} showed that one can ensure that a perturbation of $f$, $\widetilde{f}$, is $(\varepsilon, \delta)$-DP by adding a (scaled) sample $G$ from a Gaussian process prior (which uses the same kernel as $f$),
\begin{equation}
\widetilde{f} = f + {\Delta \text{c}(\delta) \over \varepsilon} G
\label{dpfunction}
\end{equation}
where DP is achieved if
\begin{equation}
\vspace{-2mm}
\text{c}(\delta) \geq \sqrt{2 \log ({2}/{\delta})}
\label{cequation}
\end{equation}
Relating this to the definition of DP, finding $\Delta$ allows us to know how much the function $f$ can change between neighbouring databases. We then choose the scale of the noise added by the randomised algorithm, $M$, to mask these changes. 

We next extend these results, from \citet{hall2013differential}, to the predictions of a GP. In the GP case we have some data (at inputs $X$ and outputs $\bm{y}$). We assume for this paper that the \emph{inputs} are non-private (e.g. people's ages), while the outputs are private (e.g. number of medications).

%\subsection{Putting a bound on the GP sensitivity}
%
The mean function of a GP posterior lies in an RKHS. We need to add a correctly scaled sample to ensure its DP release. It will become clear that the covariance function does \emph{not} need perturbation as it does not contain direct reference to the output values.% Figure \ref{kung}A demonstrates the following method, by plotting DP samples, surrounding the original GP posterior mean.

Using the notation of \citet{williams2006gaussian}, the predictive distribution from a GP at a test point $\bm{x}_*$ has mean
$
\bar{f}_* = \bm{k}_*^\top \left( K^\prime + \sigma_n^2 I \right)^{-1} \bm{y},
$
and variance
$
V[{f}_*] = k(\bm{x}_*,\bm{x}_*) -  \bm{k}_*^\top \left( K^\prime + \sigma_n^2 I \right)^{-1} \bm{k}_*,
$
where $\bar{f}_*$ is the mean of the posterior, $k(\bm{x}_*,\bm{x}_*)$ is the test point's prior variance, $\bm{k}_*$ is the covariance between the test and training points, $K^\prime$ is the Gram matrix describing the covariance between the training points, $\sigma_n^2$ is the variance of the iid noise added to each observation and $\bm{y}$ are the outputs observed values of the training data.

Ignoring any previous parameter selection, the variance does not depend on the training output values (in $\bm{y}$) so we only need to make the mean function private.

We can rewrite the above expression for the mean as the weighted sum of $n$ kernel functions,
$
\bar{f}(\bm{x}_*) = \sum_{i=1}^n {\alpha_i k(\bm{x}_i,\bm{x}_*)},
$
where $\bm{\alpha} = \left( K^\prime + \sigma_n^2 I \right)^{-1} \bm{y}$. For simplicity in the following we replace $K^\prime + \sigma_n^2 I$ with $K$, effectively combining a general kernel with a white-noise kernel.
\begin{figure}[t!]
  \centering
    \vspace{-3mm}  
    \includegraphics[width=1.05\columnwidth]{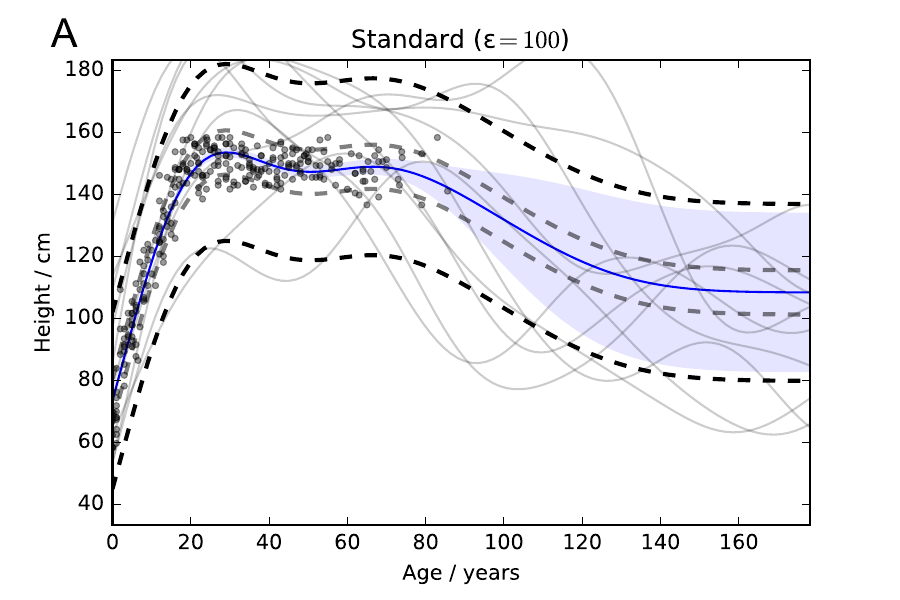}
    \vspace{-1mm}
    \includegraphics[width=1.05\columnwidth]{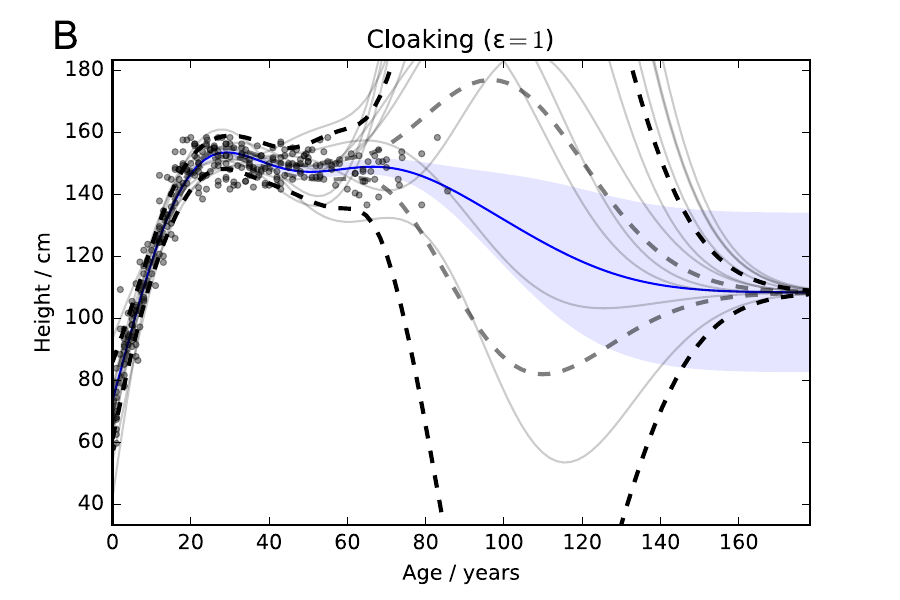}    
    \vspace{-6mm}
  \caption{Heights and ages of female !Kung San. Figure A, standard GP method. Figure B, cloaking method. One can get an intuition for the utility of the DP output by considering the example DP samples. In Figure A one can see they deviate considerably from the actual mean, even with $\varepsilon = 100$. The cloaking method, using $\varepsilon = 1$, is able to provide reasonable predictions over the domain of the training data (although not in outlier regions). Solid blue lines, posterior means of the GPs; grey lines, DP samples; Black and grey dashed lines, SE and $\frac{1}{4}$SE confidence intervals for DP noise respectively; blue area, GP posterior variance (excluding noise). $\delta=0.01$, $\Delta=100\;\text{cm}$.}
%    \vspace{}    
  \label{kung}
\end{figure}
To apply the DP algorithm described by \citet{hall2013differential} we need to find the (squared) distance in RKHS between the original and perturbed functions,
\begin{multline}
|| f_D(\bm{x_*}) - f_{D^\prime}(\bm{x_*}) ||_H^2 \\ = \Big{\langle} f_D(\bm{x_*}) - f_{D^\prime}(\bm{x_*}), f_D(\bm{x_*}) - f_{D^\prime}(\bm{x_*}) \Big{\rangle}_H.
\label{norm_squared}
\end{multline}
In \citet[section 4.1]{hall2013differential}, the vector $\bm{x}$ is identical to $\bm{x}^\prime$ with the exception of the last element $n$. In our case the inputs are identical (we are not trying to protect this part of the data). Instead it is to the values of $\bm{y}$ (and hence $\bm{\alpha}$) that we need to offer privacy.
To compute the norm in \eqref{norm_squared}, we consider the effect a difference between $\bm{y}$ and (perturbed) $\bm{y}^\prime$ has on the mean prediction function $f_D$ at $\bm{x}_*$.
\begin{multline}
f_D(\bm{x_*}) - f_{D^\prime}(\bm{x_*}) = \\ \sum_{i=1}^n \alpha_i k(\bm{x_*},\bm{x}_i) - \sum_{i=1}^n \alpha^\prime_i k(\bm{x_*},\bm{x}_i) \\ = \sum_{i=1}^n \left(\alpha_i - \alpha^\prime_i\right) k(\bm{x_*},\bm{x}_i),
\label{fun_diff}
\end{multline}
where $\bm{\alpha} = K^{-1}\bm{y}$ and (the perturbed) $\bm{\alpha}^\prime = K^{-1}\bm{y}^\prime$. In the kernel density estimation example in \citet{hall2013differential}, all but the last term in the two summations cancel as the $\alpha$ terms were absent. In our case however they remain and, generally, $\alpha_i \neq \alpha_i^\prime$. We therefore need to provide a bound on the difference between the values of $\bm{\alpha}$ and $\bm{\alpha}^\prime$.
The difference between the two vectors is,
$
\bm{\alpha} - \bm{\alpha}^\prime = K^{-1} \left(\bm{y} - \bm{y}^\prime \right).
$
As $K$ does not contain private information itself (it is dependent purely on the input and the features of the kernel) we can find a value for the bound using a specific $K$. See the supplementary material for a weaker, general, upper bound, for when the inputs are not known.

The largest change we protect against is a perturbation in \emph{one} entry of $\bm{y}$ of at most $\pm d$. Therefore we assume that all the values of $\bm{y}$ and $\bm{y}^\prime$ are equal except for the last element which differs by a value, which we will assume is a worst case of $\pm d$. Thus all the elements of $\bm{y}-\bm{y}^\prime$ are zero, except for the last, which equals $\pm d$. The result of multiplying $K^{-1}$ with the vector $\bm{y}-\bm{y}^\prime$ is the last column of $K^{-1}$ scaled by $\pm d$. Equation \ref{fun_diff} then effectively adds up the scaled last column, $\pm d [K^{-1}]_{:,n}$, but with each value scaled by the kernel's value at that point, $k(\bm{x_*},\bm{x}_i)$. We initially assume that the kernel values are bound between -1 and 1 (not unreasonable, as many kernels, such as the exponentiated quadratic, have this property, if we normalise our data). Thus a worst case result of the sum is for the positive values in the scaled column, $\pm d [K^{-1}]_{:,n}$ to be multiplied by $1$ and for the negative values to be multiplied by $-1$. Thus the largest result will be the sum of the last column's absolute ($d$-scaled) values. Finally, the \emph{use of the last column is arbitrary}, so we can bound \eqref{fun_diff} by the maximum possible sum of any column's absolute values in $K^{-1}$ (i.e. the infinity norm\footnote{The infinity norm of a symmetric square matrix is the maximum of the sums of the absolute values of the elements of rows (or columns); $\text{max}_i \sum_j |M_{ij}|$}), times $d$; i.e. $d ||K^{-1}||_\infty$.

\begin{figure*}[t!]
 \begin{minipage}[t]{0.54\textwidth}
 \vspace{0pt}
    \includegraphics[width=\textwidth]{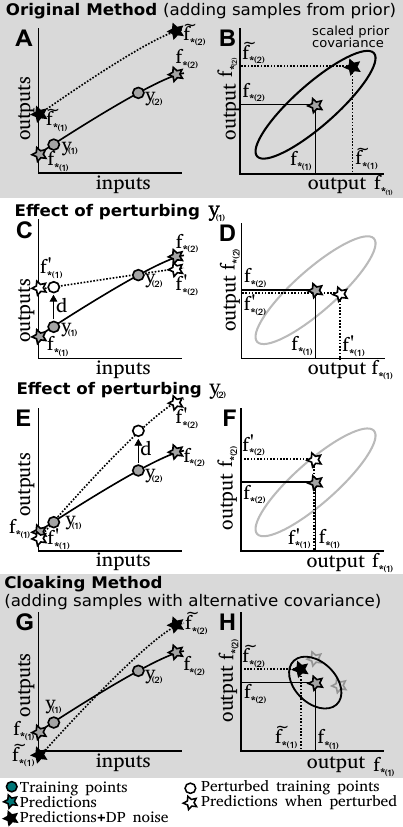}
 \end{minipage}\hfill
  \begin{minipage}[t]{0.44\textwidth}    
  \vspace{0pt}
  \caption{
  \label{main_figure}
  A illustrates the mechanism in Section \ref{noiseoutputs} in which a scaled sample from the prior has been added to the test points $\bm{f}_*$ (grey stars) to produce DP private predictions (black stars) $\widetilde{\bm{f}}_*$ using \eqref{dpfunction}. The model's long lengthscale means this moves the two test points up and down together (as they have a high covariance). Changing just one of the training points could not cause such changes in the test outputs. In B we plot the two test points against each other with the original predictions a grey star and the DP private predictions a black star. The covariance of the added DP noise is indicated with an ellipse. \\
In C we change just one of the training points, $y_{(1)}$, by adding a perturbation $d$ to it. Using \eqref{testpointchange} we can see that test point $f_{*(1)}$ increased, while $f_{*(2)}$ decreased slightly. The two test points are plotted against each other in figure D. The grey ellipse indicates the covariance of the original method's noise. The new prediction is `enclosed' by the covariance of the DP noise as the change must be indistinguishable from the DP noise. \\
In E and F we perturb the second training point, $y_2$ and plot the two test points against each other again. \\
These figures demonstrate how changing single training points can not cause perturbations like those the original method adds, in figure A. The original method, by sampling from the scaled prior, is adding more noise than we need. Instead we should be sampling from a smaller covariance which only adds noise where necessary. \\
Figures G and H illustrating this alternative covariance (ellipse in H). A DP noise sample has been added, using \eqref{vectordp}, that is as small as possible by selecting the covariance using the cloaking mechanism, while still masking the possible perturbations one could cause by changing single training points. Note that the perturbed locations from figure D and F (indicated with faint grey stars) are enclosed by the new covariance ellipse.}
  \end{minipage}
    \vspace{-3mm}    
\end{figure*}
To reduce the scale of the DP noise a little more, we very briefly consider a slightly more restrictive case, that the value of the kernel is bound between $[0, 1]$. The bound is then calculated by finding the infinity norm for the following two matrices, and taking the larger. In one $K^{-1}$ is modified so that all negative-values are ignored, in the other all values are initially negated, before the negative-values are discarded. We shall call this bound, $\text{b}(K^{-1})$. The two options described are necessary to allow us to account for the uncertainty in the sign of $\bm{y}-\bm{y}^\prime$, whose magnitude is bound by $d$ but in an unknown direction.
Returning to the calculation of the sensitivity, we can expand \eqref{fun_diff} and substitute into \eqref{norm_squared}:
\begin{multline}
\label{function_sensitivity}
||f_D(\bm{x}_*) - f_{D^\prime}(\bm{x}_*)||_H^2 \\ = \left\langle \sum_{i=1}^n (\alpha_i - \alpha^\prime_i) k(\bm{x}_*,\bm{x}_i), \sum_{i=1}^n (\alpha_i - \alpha^\prime_i) k(\bm{x}_*,\bm{x}_i) \right\rangle_H.
\end{multline}
To reiterate, we use our constraint that the chosen kernel has a range of 0 to 1, so the summations above will have a magnitude bounded by $d\, \text{b}(K^{-1})$. This means that an upper bound on the sensitivity is, 
%\begin{equation}
$
||f_D(\bm{x}_*) - f_D^\prime(\bm{x}_*)||_H^2  \leq d^2 \; \text{b}(K^{-1})^2.
$
%\label{norm_squared_2}
%\end{equation}
%
%
%
%
If two training points (with differing output values) are very close together, the mean function (and thus the bound described above) can become arbitrarily large if the Gaussian noise term $\sigma_n^2$ is small. However in reality if very nearby points have different values then the underlying system presumably has some noise involved, which we model as additional Gaussian noise. Thus the off diagonals of $K$ would remain smaller than the values on the diagonal, leading to a reasonable bound. In all the datasets examined so far, the selected Gaussian noise parameter has always been sufficiently large to avoid an excessively large bound. In general model selection for our DP GP will need to trade off between relying on single data points (i.e. low noise, causing the DP noise to be large) or relying on individual points less, due to the larger Gaussian noise term (making the non-DP prediction less accurate, but reducing the scale of the DP bound).
\subsection*{!Kung San women example}
\label{kungsan_example}
We use, as a simple demonstration, the heights and ages of 287 women from a census of the !Kung \citep{howell67}. We are interested in protecting the privacy of their heights, but we are willing to release their ages. We have set the lengthscale \emph{a priori}, to 25 years as from our prior experience of human development this is the timescale over which gradients vary.\footnote{Hyperparameters are all set \emph{a priori}, but appear precise as the data outputs were normalised to have $\mu=0$ and $\sigma=1$. Kernel variance $\sigma^2 = 7.72^2\;\text{cm}^2$, Gaussian white noise $\sigma_n^2 = 14^2$ $\text{cm}^2$, DP: $\delta = 0.01$, $\varepsilon = 50.0$, $\Delta = 100\;\text{cm}$ (enforced by rectifying all values to lie 50 cm of the mean).}
 We can find the empirical value of our sensitivity bound on the inverse covariance matrix, $\text{b}(K^{-1})$ and the value of $c(\delta)$, from \eqref{cequation}. Substituting in our given values in \eqref{dpfunction} we find that we should scale our GP samples by $28.53$. Figure \ref{kung}A shows that even with large $\varepsilon$ the DP noise overwhelms the function we want to estimate (consider the spread of DP samples in the figure). It is worth noting that, if the sensitivity of the training data had been smaller (for example count or histogram data, with $\Delta = 1$) then this method could produce usable predictions at reasonable $\varepsilon$. In the following section we find we are able to considerably reduce the scale of the DP noise by insisting that we are given the test input points \emph{a priori}.
\section{The Cloaking Method}
\label{cloaking}
The method described in the previous section is limited to low-sensitivity datasets (i.e. those for which adding a single individual would not cause much change in the posterior mean, such as histogram/count data) due to the excessive scale of the noise added. We now introduce an alternative we refer to as \emph{cloaking}, that allows a considerable reduction in the DP noise added but at the cost of needing to know the test point inputs \emph{a priori}. We approach this new method by first reasoning about the direction (across test points) noise is added by the earlier (Section \ref{noiseoutputs}) method, and comparing its effect to the effect of modifying a training point. The sensitivity in the earlier methods needed to be quite high because the noise added (sampled from the \emph{prior} covariance) is \emph{not necessarily in the direction a perturbation in a training output would cause}. 

Consider the simple case of two training and two test points, illustrated in figure \ref{main_figure}. Subfigure B illustrates (with an ellipse) the shape of the noise added to the predictions \emph{if we sample from the prior} (as in Section \ref{noiseoutputs}). Subfigures C-F illustrate changes caused by the perturbation of the training data to the predictions. The figure demonstrates that the prior does not provide the most efficient source of noise. In particular the large amount of correlated noise that is added in A and B is not necessary. Perturbations in individual training points cannot cause such correlated noise in the test outputs. To summarise; there is no perturbation in a single training point's output which could cause the predictions to move in the direction of the prior's covariance.
\vspace{-2mm}
\subsection*{Differential Privacy for vectors of GP predictions}
%
%To this end, w
From \citet[][proposition 3]{hall2013differential}: given a covariance matrix $M$ and vectors of query results (in our case GP posterior mean predictions) $\mathbf{f}_*$ and $\mathbf{f}_{*}'$ from neighbouring databases $D$ and $D'$, we define the bound,
\begin{equation}
\sup_{D \sim {D'}} ||M^{-1/2} (\mathbf{f}_* - \mathbf{f}_{*}')||_2 \leq \Delta
\label{DPdelta}
\end{equation}
$\Delta$ is a bound on the scale of the prediction change, in term of its Mahalanobis distance with respect to the added noise covariance. The algorithm provides a $(\varepsilon, \delta)$-DP output by adding scaled samples from a Gaussian distribution,
\vspace{-2mm}
\begin{equation}
\tilde{\mathbf{f}}_* = \mathbf{f}_* + \frac{\text{c}(\delta)\Delta}{\varepsilon}Z
\label{vectordp}
\end{equation}
where $Z \sim \mathcal{N}_d(0,M)$ and using function $\text{c}$ from \eqref{cequation}.
We want $M$ to have the greatest covariance in those directions most affected by changes in training points.
We are able to compute $K$, the covariance between all training points (incorporating sample variance) and $K_{*f}$ the covariance between training and test points. Given the training outputs $\mathbf{y}$, we can find the mean predictions for all test points simultaneously,
$
\mathbf{f}_* = K_{*f} K^{-1} \mathbf{y}.
$
The cloaking matrix $C = K_{*f} K^{-1}$ describes how the test points change wrt changes in training data. We use it to write the perturbed test values as
\vspace{-2mm}
\begin{equation}
\mathbf{f}_*' = \mathbf{f}_* + C \left(\mathbf{y}' - \mathbf{y} \right).
\label{testpointchange}
\end{equation}
We assume one training item $i$ has been perturbed, by at most $\pm d$: $y_i' = y_i \pm d$. As $y_i$ is the only training output value perturbed, we can see that the change in the predictions is dependent on only one column of $C$, $\mathbf{c}_i$;
$
\mathbf{f}_*' - \mathbf{f}_* = \pm d \mathbf{c}_i
$
This can be substituted into the bound on $\Delta$ in \eqref{DPdelta}. Rearranging the expression for the norm (and using $M$'s symmetry);
\begin{equation}
\begin{split}
||d M^{-1/2} \mathbf{c}_i||_2 & = (d M^{-1/2} \mathbf{c}_i)^\top (d M^{-1/2} \mathbf{c}_i) \\
% & = d \mathbf{c}_i^\top M^{-\top/2} M^{-1/2} \mathbf{c}_i d \\
 & =d^2 \mathbf{c}_i^\top M^{-1} \mathbf{c}_i
\end{split}
\end{equation}
We want to find $M$ such that the noise sample Z is small but also that $\Delta$ is minimised. A common way of describing that noise scale is to consider the determinant (generalised variance), the square root of the determinant (proportional to the volume of a confidence interval), or the log of the determinant (proportional to the differential entropy of a $k$-dimensional normal distribution (plus a constant) $\ln\left((2\pi e)^k \left|\boldsymbol\Sigma \right|\right)/2$. We use the latter but they will all have similar results.
We show in the supplementary material that the optimal $M = \sum_i{\lambda_i \mathbf{c}_i \mathbf{c}_i^\top}$ with the $\bm{\lambda}$ found using gradient descent,
$
%\frac{\partial L}{\partial \lambda_j}
{\partial L}/{\partial \lambda_j}  
%= \Tr \left( M^{-1} \bm{c}_j \bm{c}_j^\top \right) + 1
= - \bm{c}_j^\top M^{-1} \bm{c}_j + 1
$.
We return to the example of the !Kung San women data to demonstrate the improvement in privacy efficiency. Figure \ref{kung}B illustrates the results for a reasonable value of $\varepsilon=1$.\footnote{EQ kernel, $\delta=0.01$, lengthscale 25 years, Gaussian white noise 14 cm} The input domain is deliberately extended to demonstrate some features. First, where data is most concentrated the scale of the added noise is small. The effect one training point has there on the posterior prediction will be overwhelmed by its neighbours. A less intuitive finding is that the DP noise is greatest at about 110 years, far from the nearest data point. This is because the data's concentration acts like a pivot providing leverage to the outliers' effect on the posterior mean. Figure \ref{main_figure}E partly demonstrates this, with the test point $f'_{*(2)}$ being changed slightly more than the perturbation in the training point. The third observation is that the added DP noise eventually approaches zero away from the training data; the posterior mean will equal the prior mean regardless of the training data's outputs. The RMSE without DP was 6.8cm and with $(1,0.01)$-DP, 12.2cm, suggesting that this provides quite large, but practical levels of DP perturbation.
The 200 test point inputs that make up the graph's predictions are exactly those specified as part of the cloaking algorithm ($X_*$); a large number of nearby test points does not degrade the quality of the predictions as they are all very highly correlated. The noise that is already added to ensure a test point does not breach DP, is almost exactly the noise that is needed by its close neighbours.
%
%\subsection{House Prices}
%
\begin{figure}[t!]
  \centering
    \vspace{-3mm}  
    \includegraphics[width=1.02\columnwidth]{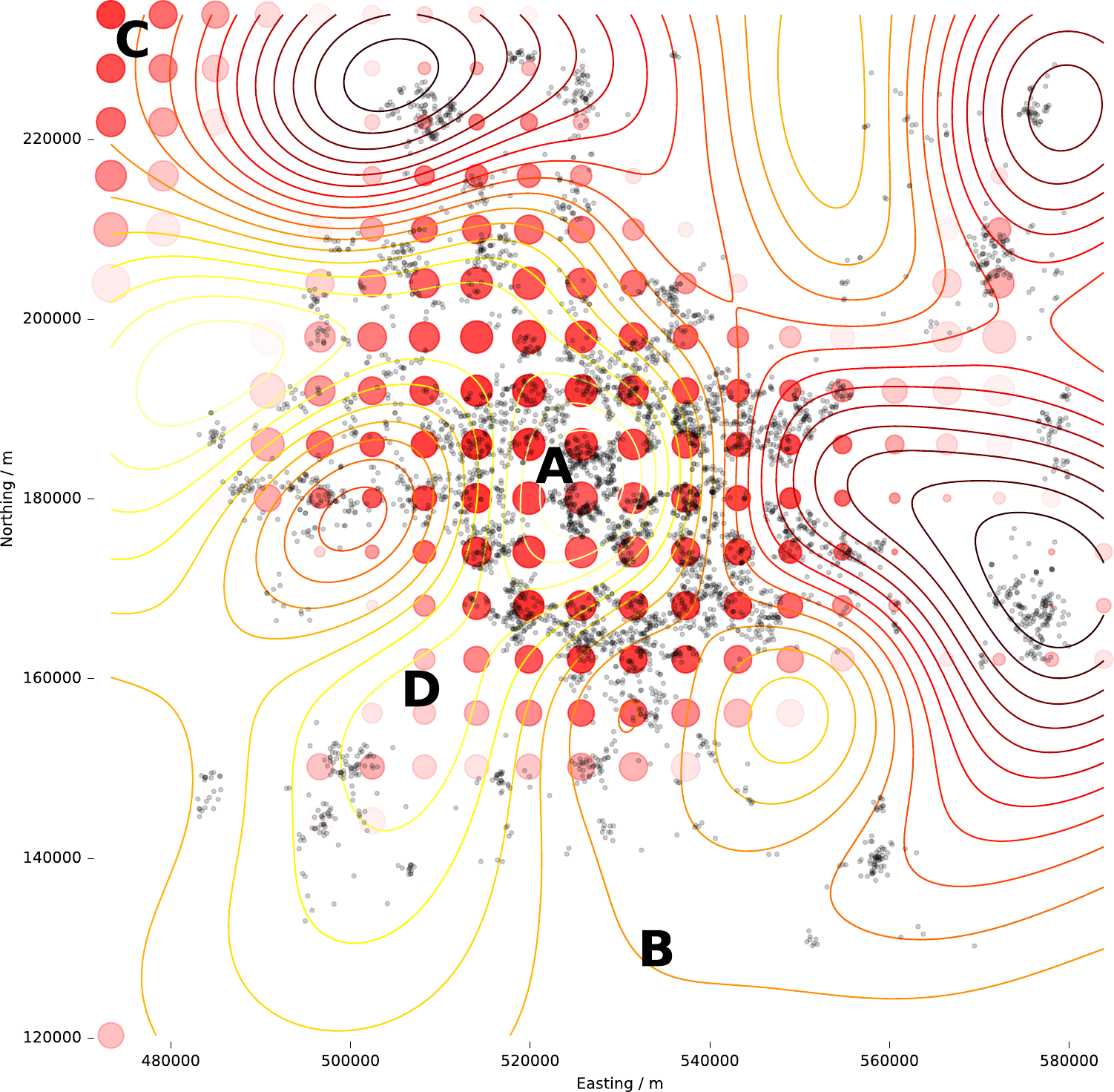}
    \vspace{-3mm}    
  \caption{4766 property prices in a $10,000\;\text{km}^2$ square around London (2013-16). Dots, property locations; circles, DP price predictions at test points. Predicted price indicated by circle area. The scale of the DP noise indicated by transparency (Opaque: no DP noise. Transparent: DP noise std is at least 40\% of $\Delta$). Non-DP predictions indicated by contours from \textsterling 215k to \textsterling 437k. $\varepsilon = 1$ and $\delta=0.01$. Areas (A) with high concentrations of training data have little DP noise, areas with few data have much more DP noise (B). Areas far from data return to GP's prior mean, and have little DP noise added (C). Interesting `bridging' effects between data concentrations cause the DP noise to remain low as the posterior is `supported' at both sides of areas with low density (e.g. D).}
    \vspace{-3mm}    
  \label{houseprices}
\end{figure}
As a further example we consider a spatial dataset of 4766 property sales since 2013 from the \citet{landregistry} in London.\footnote{Thresholded to between \textsterling100k and \textsterling500k (so the sensitivity was bounded). $\varepsilon=1$ and $\delta=0.01$, lengthscale 15 km, Gaussian variance \textsterling$^2 \text{400k}^2$.} Although this is a public dataset, one could imagine the outputs representing more private data (e.g. income or BMI). Figure \ref{houseprices} illustrates how the DP noise scale varies over the city. Training points are marked by dots. Test points are marked by the larger circles. Note how in areas where the training is concentrated the DP noise added is small, while in sparse areas the noise added is high. In the corners of the map, where there are no data, return to the GP prior's mean, and have little DP noise added.
\subsection*{Hyperparameter optimisation}
\label{hyperselection}
So far in this paper we have selected the values of the kernel hyperparameters \emph{a priori}. To demonstrate a DP method for their selection we evaluate the sum squared error (SSE) of k-fold cross-validation iterations for each hyperparameter configuration and select the optimum using the exponential mechanism (a method for selecting an item to maximise a utility, e.g. negative SSE, under DP constraints). Details of the method (including bounds on the sensitivity) can be found in the supplementary material.
\section{Results}
For comparison we binned the data (with added Laplace noise) to provide DP-predictions. Briefly, we bin the data, and add samples from the Laplace distribution with scale $\Delta f/\varepsilon$ where $\Delta f$ is the bin sensitivity (equal to the maximum change possible in height, divided by the number of data points in that bin), then fit a GP to these, now noisy, bin values. We use both a simple Exponentiated Quadratic (EQ) kernel and a kernel that models a latent function which treats the observations as integrals over each bin. We found this often does better than simple binning when the data is noisy. We did not include the standard GP method (from Section \ref{noiseoutputs}) as we found it not competitive.

%\subsection{!Kung}
For the !Kung dataset, we applied the hyperparameter selection technique to the cloaking mechanism's outputs and compared it to the results of binning. We found that hyperparameter selection, for one or two parameters, caused a reduction in RMSE that was noticeable but not impractical. Specifically we used the exponential mechanism with the negative-SSE of 100-fold Monte-Carlo cross-validation runs to select hyperparameter combinations (testing lengthscales of between 3 and 81 years, and Gaussian noise variance of between $1.1^2$ $\text{cm}^2$ and $12.7^2$ $\text{cm}^2$), which we tested against a validation set to provide the expected RMSE. We fixed the DP $\varepsilon$ to 1 for both the exponential mechanism and the cloaking and binning stages. The simple binning method (depending on bin size) had a RMSE of 23.4-50.7 cm, the integral method improved this to 14.2-20.8 cm. With no DP on parameter selection the cloaking method's RMSE would be 14.3 cm (comparable to the integral kernel's best bin size result). If we however select its hyperparameters using the exponential mechanism, the RMSE worsens to 17.4 cm. Thus there is a small, but manageable cost to the selection.
%\subsection{Citibike}
\label{citibike}
\begin{table*}[t]
  \centering
    \vspace{-3mm}    
\begin{tabular}{ l l l p{0.3cm} l p{0.3cm} l p{0.3cm} l }
\hline
 & lengthscale or bins & No DP & \multicolumn{2}{l}{$\varepsilon = 1$} & \multicolumn{2}{l}{$\varepsilon = 0.5$} & \multicolumn{2}{l}{$\varepsilon = 0.2$} \\
 \hline
 cloaking & $0.781^\circ$  & $490 \pm 14$ & $493$ & $\pm 13$ & $498$ & $\pm 13$ & $525$ & $\pm 19$ \\
          & $0.312^\circ$  & $492 \pm 15$ & $497$ & $\pm 12$ & $502$ & $\pm 17$ & $545$ & $\pm 26$ \\
          & $0.125^\circ$  & $402 \pm 7$ & $437$ & $\pm 21$ & $476$ & $\pm 17$ & $758$ & $\pm 94$ \\
          & $0.050^\circ$  & $333 \pm 11$ & $434$ & $\pm 27$ & $612$ & $\pm 78$ & $1163$ & $\pm 147$ \\
          & $0.020^\circ$  & $314 \pm 12$ & $478$ & $\pm 22$ & $854$ & $\pm 54$ & $1868$ & $\pm 106$ \\
\hline
integral binning & $10^4$ bins  & $581 \pm 5$ & $586$ & $\pm 7$ & $597$ & $\pm 12$ & $627$ & $\pm 23$ \\
                 & $6^4$ bins  & $641 \pm 6$ & $640$ & $\pm 9$ & $658$ & $\pm 17$ & $736$ & $\pm 41$ \\
                 & $3^4$ bins  & $643 \pm 6$ & $649$ & $\pm 13$ & $677$ & $\pm 22$ & $770$ & $\pm 51$ \\
 \hline
simple binning & $10^4$ bins  & $596 \pm 12$ & $1064$ & $\pm 69$ & $1927$ & $\pm 191$ & $4402$ & $\pm 434$ \\
               & $6^4$ bins  & $587 \pm 11$ & $768$ & $\pm 58$ & $1202$ & $\pm 206$ & $2373$ & $\pm 358$ \\
               & $3^4$ bins  & $550 \pm 12$ & $575$ & $\pm 24$ & $629$ & $\pm 58$ & $809$ & $\pm 110$ \\
\hline
\end{tabular}
    \vspace{-1mm}  
\caption{RMSE (in seconds, averaged over 30-fold X-validation, $\pm$ 95\% CIs) for DP predictions of Citi Bike journey durations. Five lengthscales (in degrees latitude/longitude) and three bin resolutions were tested for the cloaking and binning experiments respectively. The cloaking method, with the right lengthscale, makes more accurate predictions than either of the binning methods. As we increase $\varepsilon$, cloaking needs longer lengthscales to remain competitive as this allows the predictions to `average' over more training data.}
\label{citibike_table}
    \vspace{-3mm}  
\end{table*}
Using data from the New York City bike sharing scheme, \citet{citibike}\footnote{163,000 subscribers, 600 stations located in a box bounded between latitudes $40.6794^\circ$ and $40.7872^\circ$, and longitudes $-74.0171^\circ$ and $-73.9299^\circ$. Unlike the house-price data we kept the locations in these global coordinates. Each fold of the Monte Carlo cross validation sampled 5000 rows from the 1,460,317 journeys in June 2016.} we predict journey time, given the latitude and longitude of the start and finish stations. The 4d Exponentiated Quadratic (EQ) kernel had lengthscales of between 0.02 and 0.781 degrees (latitude or longitude, equivalent to roughly 1.8 km to 75 km. $\sigma^2 = 1581^2\;\text{s}^2$, $l = 0.05^\circ$, $\sigma_n = \text{1605}^2$ $\text{s}^2$. Durations were thresholded to a maximum of 2000 s.

We tested values of $\varepsilon$ between 0.2 and 1.0 (with $\delta$ fixed at 0.01) and with DP disabled. Monte Carlo cross-validation was used to make predictions using the DP framework (4900 training, 100 test journeys). For comparison we binned the training data into between 81 and 10,000 bins, then computed DP means for these bins. These DP values were used as predictions for the test points that fell within that bin (those bins without training data were set to the population mean). Table \ref{citibike_table} summarises the experimental results.
The new cloaking function achieves the lowest RMSE, unless both $\varepsilon$ and the lengthscales are small.
With no DP the short-lengthscale cloaking method provides the most accurate predictions, as this is not affected by the binning and is capable of describing the most detail in the spatial structure of the dataset. The simple binning becomes less accurate with more bins, probably due to low occupancy rate (in a random training example, with 10,000 bins, only 11\% were occupied, and of those 40\% only had one point) and a form of overfitting. As $(1,0.01)$-DP noise is added the simple-binning degrades quickly due to high DP noise added to low-occupancy bins. \citet{xu2013differentially} also describes a similar phenomenon. Predictions using the GP with an integral kernel fitted to these DP bin counts appears to provide some robustness to the addition of DP noise.
As $\varepsilon$ is further reduced, the cloaking method does better at longer lengthscales which allow more averaging over the training data. Simple binning becomes increasingly compromised by the DP noise.

\section{Discussion}
The cloaking method performs well, providing reasonably constrained levels of DP noise for realistic levels of privacy and provides intuitive features such as less DP-noise in those areas with the greatest concentration of training data. The earlier method, described in Section \ref{noiseoutputs}, required much more DP perturbation.
For the cloaking method the lengthscale provides a powerful way of trading off precision in modelling spatial structure with the costs of increasing DP-noise. We could exploit this effect by using non-stationary lengthscales \citep[e.g.][]{snoek2014input, herlands2015scalable}, incorporating fine lengthscales where data is common and expansive scales where data is rarefied. This could lead to DP noise which remains largely constant across the feature space.
To further reduce the DP noise, we could manipulate the sample noise for individual output values. For the initial method, by adjusting the sample noise for individual elements we can control the infinity-norm. For the cloaking method, outlying training points, around the `edge' of a cluster, could have their sample noise increased.
One important issue is how to make DP GP predictions if we want to protect the values of the training \emph{inputs}. This could be approached by considering a bound on the inverse-covariance function, a suggestion is provided in the supplementary material.

Future work is also needed to address how one optimises the hyperparameters of these models in a DP way. The method described in Section \ref{hyperselection} works but is far from optimal. It may be possible to extend methods that use DP Bayesian optimisation to estimate values for hyperparameters \citep{kusner2015differentially}, or approximate the likelihood function to work with the method described in \citet{han2014differentially}. It would also be interesting to investigate how to modify the objective function to incorporate the cost of DP noise.

The actual method for releasing the corrupted mean function for the output-noise methods has not been discussed. Options include releasing a set of samples from the mean and covariance functions (a necessary step for the cloaking method, as the test-points need specifying in advance), or providing a server which responds with predictions given arbitrary input queries, sampling from the same Gaussian (for the cloaking method, querying new test points could be achieved by conditioning on the outputs given previously).
The examples given here all use the EQ kernel but the cloaking method works with arbitrarily complex kernel structures and it has no requirement that the covariance function be stationary. GP classification is also an obvious next step.

Finally, in unpublished work, we have found evidence that the use of inducing inputs can significantly reduce the sensitivity, and thus DP noise required by reducing the predictions' dependency on outliers. Future work should be undertaken to investigate the potential for further reducing the DP noise through the use of inducing inputs.

We have presented novel methods for combining DP and GPs. In the longer term we believe a comprehensive set of methodologies could be developed to enhance their applicability in privacy preserving learning. We have applied DP for functions to GPs and given a set of known test points we were able to massively reduce the scale of perturbation for these points by considering the structure of the perturbation sensitivity across these points. In particular we found that the cloaking method performed considerably more accurately than the binning alternatives.

\subsection*{Acknowledgements}
This work has been supported by the Engineering and Physical Research Council (EPSRC) Research Project EP/N014162/1.

\bibliographystyle{plainnat}
\bibliography{refs}

\clearpage
\section*{Supplementary Material}

\subsection*{Deriving and optimising the cloaking variables}
When solving for $M$, we put an arbitrary (positive) bound on $\Delta$ of 1, as any scaling of $\Delta$ caused by manipulating $M$, will scale the determinant of $M$ by the inverse amount.\footnote{For example if we write a new $\Delta'$ with $M$ as $mM$, we see that we can take out the $m$ term from $\Delta'$'s inequality, leaving $\Delta' = m^{-1/2} \Delta$. When we scale Z, which has covariance $mM$ by this new value of $\Delta'$ the covariance of the scaled Z equals $(\Delta')^2 mM = (m^{-1/2} \Delta)^2 m M = \Delta^2 M$, the magnitude change cancels, so any positive value of $\Delta$ (e.g. 1) will suffice for the optimisation. Also: in the following we ignore $d$ and reintroduce it at the end of the derivation by defining $\Delta = d$ instead of it equalling 1.}

We will express our problem as trying to \emph{maximise} the entropy of a $k$-dimensional Gaussian distribution with covariance $P = M^{-1}$;

\begin{quotation}
Maximise $\ln\left(\left| P \right| \right)$,
subject to $n$ constraints,
$0 \leq \mathbf{c}_i^\top P \mathbf{c}_i \leq 1$.
\end{quotation}

%We will see that the way the solution is constructed means that $P$ is positive semidefinite, thus we will ignore the lower bound for now.
%Remove?
%\footnote{as for any positive semidefinite matrix, $\mathbf{c}_i^\top P \mathbf{c}_i \geq 0$, for any vector $\mathbf{c}_i$}.

Considering just the upper bounds and expressing this as a constraint satisfaction problem using Lagrange multipliers we have 
$
L = \ln \left| P \right| + \sum_i{\lambda_i (1 - \mathbf{c}_i^\top P \mathbf{c}_i)}.
$
Differentiating (and setting to zero),
$
\frac{\partial{L}}{\partial{P}} = P^{-1} - \sum_i{\lambda_i \mathbf{c}_i \mathbf{c}_i^\top} = 0.
$
We also have the slackness conditions (for all $i$), $\lambda_i (\mathbf{c}_i^\top P \mathbf{c}_i - 1) = 0$ and we also require that $\lambda_i \geq 0$.
Rearranging the derivative, we have,
$
P^{-1} = \sum_i{\lambda_i \mathbf{c}_i \mathbf{c}_i^\top}.
$
Note that as $\lambda_i \geq 0$, $P^{-1}$ is positive semi-definite (psd),\footnote{The summation is of a list of positive semi-definite rank-one matrices. One can see that such a sum is also positive semi-definite (to see this, consider distributing $\bm{z}^\top$ and $\bm{z}$ over the summation).} thus the initial lower bound (that $\mathbf{c}_i^\top P \mathbf{c}_i \geq 0$) is met. The upper bound (that $\mathbf{c}_i^\top P \mathbf{c}_i \leq 1$) is achieved if the $\lambda_i$ are correctly chosen, such that the Lagrange and slackness conditions are met.

We now must maximise the expression for $L$ wrt $\lambda_j$. To assist with this we rewrite our expression for 
$
P^{-1} = \sum_i{\lambda_i \mathbf{c}_i \mathbf{c}_i^\top} = C \Lambda C^\top,
$
where $C = \left[\bm{c}_1, \bm{c}_2 ... \bm{c}_n\right] = K_{*f} K^{-1}$ and $\Lambda$ is a diagonal matrix of the values of $\lambda_i$.
The summation in the expression for $L$, $\sum_i{\lambda_i (1 - \mathbf{c}_i^\top P \mathbf{c}_i)}$ can be written as $\Tr \left(\Lambda - C^\top P C \Lambda \right)$. Substituting in our definition of $P$, we can write the summation as:
$\Tr \left(\Lambda - C^\top (C \Lambda C^\top)^{-1} C \Lambda \right) = \Tr \left(\Lambda - C^\top C^{-\top} \Lambda^{-1} C^{-1} C \Lambda \right)$. Assuming $C$ is invertible, the summation becomes $\Tr \left(\Lambda - \Lambda^{-1} \Lambda \right)$. Differentiating this wrt $\lambda_j$ equals one.
We can use this result to find the gradient of $L$ wrt $\lambda_j$:
\begin{equation}
\begin{split}
\frac{\partial L}{\partial \lambda_j} &= \frac{\partial \ln{|P|}}{\partial \lambda_j} + \frac{\partial}{\partial \lambda_j} \sum_i{\lambda_i (1 - \mathbf{c}_i^\top P \mathbf{c}_i)}\\
&= \Tr \left( P^{-1} \frac{\partial P}{\partial \lambda_j} \right) + 1\\
&= \Tr \left( -P^{-1} P \bm{c}_j \bm{c}_j^\top P \right) + 1\\
&= - \Tr \left( \bm{c}_j \bm{c}_j^\top M^{-1} \right) + 1\\
&= - \bm{c}_j^\top M^{-1} \bm{c}_j + 1
\end{split}
\end{equation}
This can be solved using a gradient descent method, to give us those values of $\lambda_i$ which minimise $\log |M|$ while ensuring $\Delta \leq 1$.

\subsection*{Citi Bike duration distribution}

\begin{figure}[t!]
  \centering
    \includegraphics[width=0.6\columnwidth]{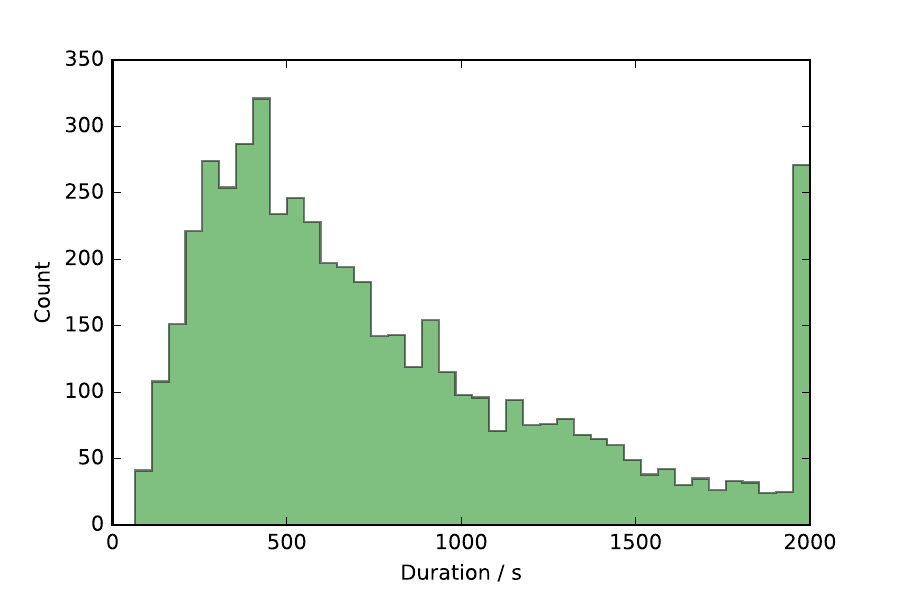}
  \caption{The duration of 5000 Citi Bike journeys. Note the effect caused by thresholding at 2000 s.}
  \label{citibike_hist}
\end{figure}

Figure \ref{citibike_hist} illustrates the distribution of journey times and the effect of thresholding.

\subsection*{Algorithms}
\newcommand*\Let[2]{\State #1 $\gets$ #2}
Algorithm \ref{Cloaking_algorithm} describes the Cloaking method. Algorithm \ref{simple_algorithm} describes the earlier method, from Section \ref{noiseoutputs}.
\begin{algorithm*}
  \newcommand{\Model}{\mathcal{M}}
  \caption{Using the initial DP algorithm \label{simple_algorithm}}
  \begin{algorithmic}[1]
    \Require{$\Model$, the GP model (the kernel, hyperparameters and training inputs and normalised outputs)}
    \Require{$X_{*} \in R^{P \times D}$, (the matrix of test inputs)}
    \Require{$d > 0$, data sensitivity (maximum change possible) \& $\varepsilon>0, \delta \geq 0$, the DP parameters}
    \Statex
    \Function{DifferentiallyPrivateRegression}{$X_{*}$, $M$, $d$, $\varepsilon$, $\delta$}
      \Let{$K$}{$\Model\textsc{.get\_K}$} \Comment{covariance between training points}      
      \Let{$\Delta$}{$d^2 \textsc{b}(K^{-1})^2$}
      \Let{$\bm{f}_*, \sigma_*^2$}{$\Model\textsc{.get\_predictions}(X_*)$} \Comment{Calculate non-DP predictions}
      \Let{$\tilde{\bm{f}}_*$}{$\bm{f}_* + (\Delta d \textsc{c}(\delta) / \varepsilon) \bm{z}$} \Comment{$\bm{z} \sim \mathcal{N}(0, K)$}
      \State \Return{$\tilde{\bm{f}}_*$, $\sigma_*^2$}
    \EndFunction
    
    \Statex    
    \Function{b}{$K^{-1}$}
    \State \Return{$\text{max}(|-[K_A^{-1}]^{-}|_\infty,|[K_A^{-1}]^{+}|_\infty)$}
    \EndFunction

    \Statex    
    \Function{c}{$\delta$}
    \State \Return{$\sqrt{2 log(2/\delta)}$}
    \EndFunction    
  \end{algorithmic}
\end{algorithm*}

\begin{algorithm*}
  \newcommand{\Model}{\mathcal{M}}
  \newcommand{\grad}{\frac{dL}{d\bm{\lambda}}}
  \newcommand{\lam}{\bm{\lambda}}
  \caption{Using the cloaking algorithm \label{Cloaking_algorithm}}
  \begin{algorithmic}[1]
    \Require{$\Model$, the GP model (the kernel, hyperparameters and training inputs and normalised\textsuperscript{*} outputs)}
    \Require{$X_{*} \in R^{P \times D}$, (the matrix of test inputs)}
    \Require{$d > 0$, data sensitivity (maximum change possible) \& $\varepsilon>0, \delta \geq 0$, the DP parameters}
    \Statex
    \Function{DifferentiallyPrivateCloakingRegression}{$X_{*}$, $M$, $d$, $\varepsilon$, $\delta$}
      \Let{$C$}{$\Model\textsc{.get\_C}(X_{*})$} \Comment{Compute the value of the cloaking matrix ($K_{*f} K^{-1}$)}
      \Let{$\lam$}{$\textsc{findLambdas}(C)$}
      \Let{$M$}{$\textsc{calcM}(\lam,C)$} \Comment{Calculate the DP noise covariance matrix}
      \Let{$\Delta$}{$\textsc{calcDelta}(\lam,C)$\textsuperscript{\dag}}
      \Let{$c$}{$\sqrt{2 log \frac{2}{\delta}}$}
      \Let{$\bm{y}_*, \sigma_*^2$}{$\Model\textsc{.get\_predictions}(X_*)$} \Comment{Calculate non-DP predictions}
%      \State{$\bm{z} \sim \mathcal{N}(0, M)$}
      \Let{$\tilde{\bm{y}}_*$}{$\bm{y}_* + (\Delta d c / \varepsilon) \bm{z}$} \Comment{$\bm{z} \sim \mathcal{N}(0, M)$}
      \State \Return{$\tilde{\bm{y}}_*$, $\sigma_*^2$}
    \EndFunction
    
    \Statex    
    \Function{$\Model\text{.get\_C}$}{$X_*$}
        \State From $\Model$ compute $K_{*f}$ and $K^{-1}$ \Comment{Compute covariances between training and test points}
        \Let{$C$}{$K_{*f} K^{-1}$}
    \State \Return{$C$}
    \EndFunction
    
    \Statex    
    \Function{findLambdas}{$C$}
        \Let{$\bm{\lambda}$}{$\textsc{Uniform}(0.1,0.9)$} \Comment{Initialise randomly\textsuperscript{\ddag}}
        \Let{$\alpha$}{$0.05$} \Comment{Learning rate}
        \Do
            \Let{$\grad$}{$\textsc{CalcGradient}(\lam,C)$}
            \Let{$\Delta_{\lam}$}{$-\grad \alpha$}
            \Let{$\lam$}{$\lam + \Delta_{\lam}$}
        \doWhile{$\Delta_{\lam} > 10^{-5}$}
    \State \Return{$\lam$}
    \EndFunction
    
    \Statex    
    \Function{calcGradient}{$\lam, C$}
        \Let{M}{$\textsc{calcM}(\lam,C)$}
        \For {$0 \leq j < N$} \Comment{N, number of columns in cloaking matrix, C.}
        \Let{$[\grad]_j$}{$-\Tr \left(M^{\scalebox{0.8}{+}} C_{:j} C_{:j}^\top\right) +1$}
        \EndFor
    \State \Return{$\grad$}
    \EndFunction
    
    \Statex    
    \Function{calcM}{$\lam, C$}
        \Let{$M$}{$\sum_i{\lambda_i C_{:i} C_{:i}^\top}$}
    \State \Return{$M$}
    \EndFunction
    
    \Statex    
    \Function{calcDelta}{$\lam, C$}
        \Let{$M$}{$\textsc{calcM}(\lam,C)$}
        \Let{$\Delta$}{$\text{\textbf{max}}_j$ $C_{:j}^\top M^{\scalebox{0.8}{+}} C_{:j}$}
    \State \Return{$\Delta$}
    \EndFunction
  \end{algorithmic}
  
  \footnotesize{\textsuperscript{*}We assume the user will handle normalisation.}
  
  \footnotesize{\textsuperscript{\dag}Although we should have optimised $M$ such that $\Delta \leq 1$, it may not have completely converged, so we compute the $\Delta$ bound for the value of $M$ we have actually achieved.}
  
  \footnotesize{\textsuperscript{\ddag}We have found that occasionally the algorithm fails to converge. To confirm convergence we have found it useful to reinitialise and run the algorithm a few times.}
\end{algorithm*}

\subsection*{Hyperparameter selection}
%\subsubsection*{Background}
So far in this paper we have selected the values of the kernel hyperparameters \emph{a priori}. Normally one may maximise the marginal likelihood to select the values or potentially integrates over the hyperparameters. In differential privacy we must take care when using private data to make these choices. Previous work exists to perform this selection, for example \citet{kusner2015differentially} describes a method for performing differentially private Bayesian optimisation, however their method assumes the training data is not private.
%from Kusner '15: For the case of machine learning hyper-parameter tuning our results are designed to guarantee privacy of the validation set only (it is equivalent to guarantee that the training set is never allowed to change).  To simultaneously protect the privacy of the training set it may be possible to use techniques similar to the training stability results of Chaudhuri & Vinterbo (2013). Training stability could be guaranteed, for example, by assuming an additional training set kernel that bounds the effect of altering the training set on f. We leave developing these guarantees for future work. 
\citet{kusner2015differentially} do suggest that the work of \citet{chaudhuri2013stability} may allow Bayesian Optimisation to work in the situation in which the training data also needs to be private.

%\subsubsection*{Bounds}
We decided instead that, due to the low-dimensionality of many hyperparameter problems, a simple grid search, combined with the exponential mechanism may allow the selection of an acceptable set of hyperparameters. For the utility function we considered using the log marginal likelihood, with additional noise in the data-fit term to capture the effect of the DP noise. However for simplicity in estimating the bound and to avoid overfitting we simply used the sum square error (SSE) over a series of $K$-fold cross-validation runs, which for a given fold is $\sum_{i=1}^{N} \left( y_{*i} - y_{\text{t}i} \right)^2$, with predictions $\bm{y}_*$ and test values $\bm{y}_{\text{t}}$.

Before proceeding we need to compute a bound on the sensitivity of the SSE. To briefly recap, the DP assumption is that one data point has been perturbed by at most $d$. We need to bound the effect of this perturbation on the SSE. First we realise that this data point will only be in the \emph{test set} in one of the $K$ folds. In the remaining folds it will be in the training data.

If the perturbed data point is in the training data ($\bm{y}$), then we can compute the sensitivity of the SSE. The perturbation this would cause to the predictions ($\bm{y}_*$) is described using standard GP regression (and the cloaking matrix). Specifically a change of $d$ in training point $j$ will cause a $d\bm{c}_{jk}$ change in the test point predictions, where $\bm{c}_{jk}$ is the $j$th column of the cloaking matrix for the $k$th fold.

To compute the perturbation caused by the change in the training data, we note that the SSE is effectively the square of the euclidean distance between the prediction and the test data. We are moving the prediction by $d\bm{c}_{jk}$. The largest effect that this movement of the prediction point could have on the distance between prediction and test locations is if it moves the prediction in the opposite direction to the test points. Thus it can increase (or decrease) the distance between the test and predictions by the largest length of $d\bm{c}_{jk}$ over training points. Hence for one of the folds, the largest change in the SSE is $d^2 \max_j |\bm{c}_{jk}|_2^2$.

If the perturbed data point, $j$, was in the test data then the SSE will change by $\left( y_{*j} + d - y_{\text{t}j} \right)^2 - \left( y_{*j} - y_{\text{t}j} \right)^2 = d^2 + 2d (y_{*j} - y_{\text{t}j})$. The last part of the expression (the error in the prediction for point $j$) is unbounded. To allow us to constrain the sensitivity we enforce a completely arbitrary bound of being no larger than $\pm 4d$, thresholding the value if it exceeds this. Thus a bound on the effect of the perturbation is $d^2 + 2d \times 4d = d^2 + 8d^2 = 9d^2$.

The SSE of each fold is added together to give an overall SSE for the cross-validation exercise. We sum the $K-1$ largest sensitivities and add $9d^2$ to account for the effect of the single fold in which the perturbing data point, $j$, will be in the test set. The perturbation could have been in the test data in any of the folds. We assumed it was in the fold with the smallest training-data sensitivity to allow us to make the result a lower bound on the sensitivity of the SSE to the perturbation. If it had been in any other fold the sensitivity would have been less (as more sensitivity would be contributed by the sum over the training sensitivities). Thus the sensitivity of the SSE over the $K$ folds is; $9d^2 + \sum_{k=1}^{K-1} d^2 \max_j |\bm{c}_{jk}|_2^2$ (where the k folds are ordered by decreasing sensitivity)

We compute the SSE and the SSE's sensitivity for each of the hyperparameter combinations we want to test. We then use the computed sensitivity bound with the exponential mechanism to select the hyperparameters. To summarise, to use the exponential mechanism one evaluates the utility $u(x,r)$ for a given database $x$ and for elements $r$, from a range. One also computes the sensitivity of this utility function by picking the highest sensitivity of any of the utilities; in this case each utility corresponds to a negative SSE, and each sensitivity corresponds to the sum described above.

$$\Delta_u \triangleq \underset{r \in R}{\max}\; \underset{x,y}{\max}\; |u(x,r)-u(y,r)|$$ (where $x$ and $y$ are neighbouring databases).

The exponential mechanism selects an element $r$ with probability proportional to:

$$\exp \left( \frac{\varepsilon u(x,r)}{2 \Delta_u} \right)$$

Note that for a given privacy budget, some will need to be expended on this selection problem, and the rest expended on the actual regression.

\subsubsection*{Effect of privacy on optimum hyperparameter selection}
\begin{figure}[t!]
  \centering
    \vspace{-3mm}
    \includegraphics[width=\columnwidth]{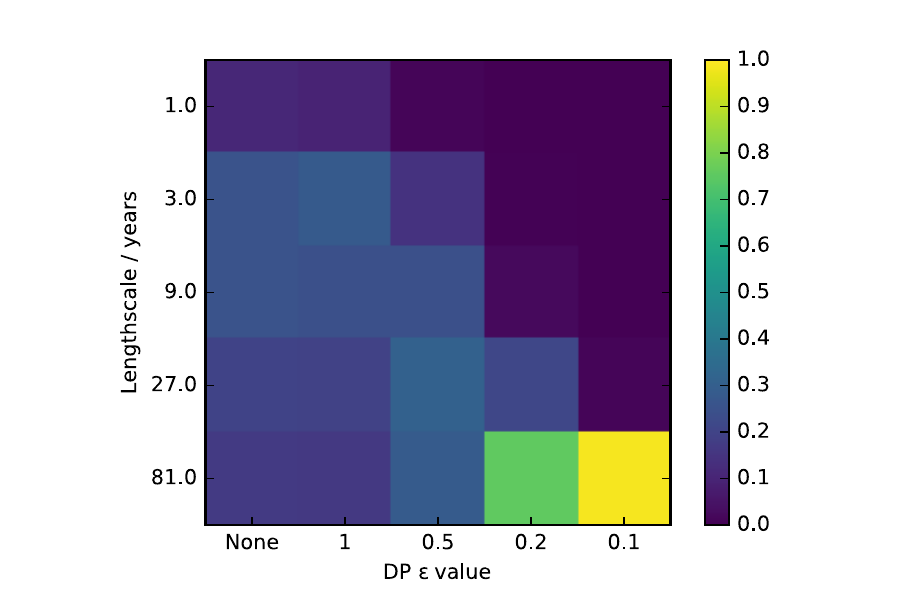}
    \vspace{-3mm}
  \caption{Effect of varying the differential privacy parameter, $\varepsilon$, on the likelihood of selecting each lengthscale. Colour indicates probability of parameter selection. With low privacy, a short lengthscale is appropriate which allows the GP to describe details in the data. With high privacy, a longer lengthscale is required, which will average over large numbers of individual data points.}
  \vspace{-3mm}
  \label{effectofepsonls}
\end{figure}

A final interesting result is in the effect of the level of privacy in the regression stage on the selection of the lengthscale. This is demonstrated in the distribution of probabilities over the lengthscales when we adjust $\varepsilon$. Figure \ref{effectofepsonls} demonstrates this effect. Each column is for a different level of privacy (from none to high) and each tile shows the probability of selecting that lengthscale. For low privacy, short lengthscales are acceptable, but as the privacy increases, averaging over more data allows us to give more accurate answers.

\subsection*{Privacy on the training inputs}
To release the mean function such that the training inputs remain private, we need a general bound on the infinity norm of the covariance function, that does not depend explicitly on the values of $X$.

\citet{varah1975lower} show that if $J$ is \emph{strictly diagonally dominant}\footnote{A matrix, $J$, is strictly diagonally dominant if $\Delta_i(J) > 0$ for all $1 \leq i \leq n$.} then:
$$||J^{-1}||_\infty \leq \max\limits_{1 \leq i \leq n} \frac{1}{\Delta_i(J)} = b(J)$$

where we have defined this bound as $b(J^{-1})$. We also define $\Delta_i(J) = |J_{ii}| - \sum_{j \neq i} |J_{ij}|$, i.e. the sum of the off diagonal elements in row $i$ subtracted from the diagonal element. 

So if $K$ is strictly diagonally dominant (which is achieved if the inputs are sufficiently far apart, and/or if sufficient uncertainty exists on the outputs), then we have a bound on the sums of its rows. The above bound means that,
\begin{equation}
  \sum_{i=1}^n \alpha_i - \alpha^\prime_i \leq \Delta_y b(J^{-1})
\end{equation}

To ensure sufficient distance between inputs, we could use inducing variables, which can be arbitrarily placed, so that the above constraints on the covariance matrix are observed.

\subsection*{Integral Kernel}

We used a custom kernel in the Citi Bike comparison to make predictions over the binned DP-noisy dataset. In summary the observations were considered to be the integrals over each bin (provided by multiplying the noisy means by the sizes of the bins). The predictions were made at points on the latent function being integrated. This allowed us to make predictions from the binned data in a principled manner.

Considering just one dimension. To compute the covariance for the integrated function, $F(\cdot)$, we integrate the original EQ kernel, $k_{ff}(\cdot,\cdot)$, over both its input values,
$$
k_{FF}((s,t), (s^\prime, t^\prime)) = \alpha\; \int_s^t \int_{s^\prime}^{t^\prime} k_{ff}(u,u^\prime)\; \text{d}u^\prime \text{d}u,
$$
so that given two pairs of input locations, corresponding to two definite integrals, we can compute the covariance between the two.

Similarly we can calculate the cross-covariance $k_{Ff}$ between $F$ and $f$. Both $k_{FF}$ and $k_{Ff}$ can be extended to multiple dimensions. Each kernel function contains a unique lengthscale parameter, with the bracketed kernel subscript index indicating these differences. We can express the new kernel as the product of our one dimensional kernels:
$$
k_{FF} = \prod_i k_{FF(i)}((s_i,t_i),(s^\prime_i,t^\prime_i)),
$$
with the cross covariance given by
$$
k_{Ff} = \prod_i k_{Ff(i)}((s_i,t_i),(s^\prime_i,t^\prime_i)).
$$
The above expressions can then be used to make predictions of the latent function $f$ given observations of its definite integrals.
\end{document}